\documentclass[letterpaper, 10 pt, conference]{IEEEtran}
\IEEEoverridecommandlockouts
\usepackage{cite}
\usepackage{amsmath,amssymb,amsfonts}
\usepackage{blindtext}
\usepackage{algpseudocode}
\usepackage{algorithm}
\usepackage{float}
\usepackage{subfig}
\usepackage{graphicx}
\usepackage{optidef}
\usepackage{textcomp}
\usepackage{xcolor}
\usepackage{optidef}
\usepackage[font=footnotesize]{caption}
\def\BibTeX{{\rm B\kern-.05em{\sc i\kern-.025em b}\kern-.08em
    T\kern-.1667em\lower.7ex\hbox{E}\kern-.125emX}}

\begin{document}

\title{\bf\Large Deformable Linear Object Surface Placement \\
Using Elastica Planning and Local Shape Control}

\author{I. Grinberg$^1$ and A. Levin$^2$ and E. D. Rimon 
\thanks{This work receives funding from uropean Union's research and innovation program under grant agreement no. $101070600$, project SoftEnable.~Authors $1 $ and~$2$ made equal contributions.
Dept. of ME, Technion, Israel.}
\vspace{-.3in}
}

\maketitle

\begin{abstract}
Manipulation of deformable linear objects ({\small DLO}s) in constrained environments is a challenging task. 
This paper describes a two-layered approach for placing {\small DLO}s  on a flat surface using a single robot hand. The high-level layer is a novel {\small DLO} surface placement method based on Euler's elastica solutions. During this process one
{\small DLO} 
endpoint is manipulated by the robot gripper while a variable interior point of the {\small DLO} serves as the start point of the  portion aligned with the placement surface. The low-level layer forms a pipeline controller. The controller estimates the {\small DLO} current shape using a Residual Neural Network (ResNet) and uses low-level feedback to ensure task execution in the presence of 
modeling and placement errors. 
The resulting {\small DLO} placement approach can recover from states where the high-level manipulation planner has failed as required by practical robot manipulation systems.  The 
{\small DLO} placement 
approach is demonstrated with simulations and experiments that use silicon mock-up objects prepared for fresh food applications.
\end{abstract}


\section*{\textbf{ I. Introduction}}
\vspace{-.06in}

\begin{figure}
\centerline{\includegraphics[scale=0.33]{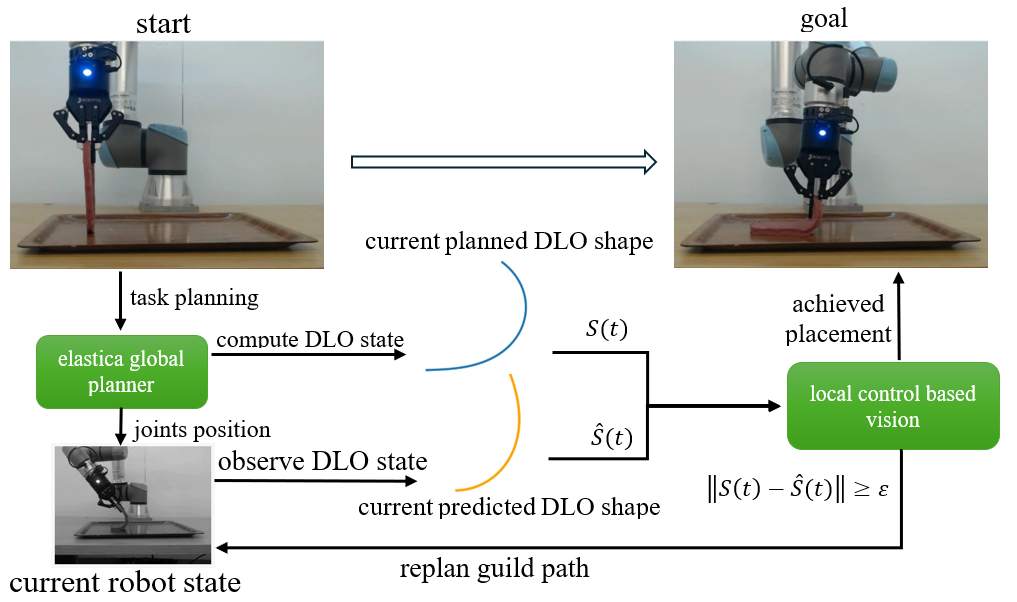}}
\caption{A single robot arm has to place a~strip-like fresh food item whose median axis is modeled as a {\footnotesize DLO} on a tray in a~stable and non-self-intersecting manner.
The {\footnotesize DLO} placement combines high-level planning with low-level feedback controller: $S(t)$ is the {\footnotesize DLO} high-level planned shape, $\hat{S}(t)$ is the  ResNet measurement {\footnotesize DLO} observed shape. {\footnotesize $||S(t) \!-\! \hat{S}(t)||$} is the {\footnotesize DLO} shape estimation error fed to the low-level controller.}
\label{framework.fig}
\vspace{-.25in}
\end{figure}

\noindent This paper considers robotic~\mbox{manipulation}~of~\mbox{deformable}~\mbox{linear} objects, called {\small DLOs}.~In~these~\mbox{problems},~one~or~two~robot~hands apply endpoint forces and torques that together with external influences such as gravity and environmental constraints affect the {\small DLO} shape during manipulation. This paper is motivated by SoftEnable project~\cite{SoftEnable} where {\small DLOs}~are~used~to model strip like fresh food items (Fig.~\ref{framework.fig}). Other robotic applications include parts assembly~[10], rope routing and untangling~\cite{bretl_multiple,cable_rss22} surgical suturing~\cite{jackson&cavusogl} and agricultural robotics~\cite{adaptive_LDO22}. {\small DARPA}'s Plug-Task~\cite{chang&padir}, EU IntelliMan~\cite{IntelliMan}
and SoftEnable~\cite{SoftEnable} and ICRA workshops~\cite{zhu_survey22,deform_23,deform_24} are all dedicated to soft and deformable object handling.\\
\indent In previous work~\cite{Levin&Grinberg&Rimon&Shapiro}, we described a dual arm {\small DLO} steering scheme in the presence of obstacles. By exploiting Euler's elastica solutions for {\small DLO} equilibrium shapes in two-dimensions, \cite{Levin&Grinberg&Rimon&Shapiro} plans  {\small DLO} steering paths~in~two-dimensions and in semi-spatial manner in three-dimensions.
However, the {\small DLO} was not allowed to contact features of the environment as such contacts apply reaction forces that affect the {\small DLO} shape.
This paper considers manipulating {\small DLOs} by a single robot arm rather than dual arms. The single arm holds the {\small DLO} at one end while placing it on a tray or other flat surface in a stable non-self-intersecting and non-sliding manner (Fig.~\ref{framework.fig}). Friction effects between the {\small DLO} and tray and between tray and table are also taken into account during placement. High-level planning of the robot hand movement uses the 
elastica shape parameters. A low-level controller uses a~Residual Neural Network (ResNet) to ensure task execution in the presence of {\small DLO}  modeling and placement errors (Fig.~\ref{framework.fig}). 



{\bf Related work:} The equilibrium shapes of {\small DLOs} held under endpoint forces and moments in two-dimensionas satisfy {\em Euler-Bernoulli's bending law,} where the internal moment at each point 
is proportional to the {\small DLO} curvature at this point. See Wakamatsu~\cite{wakamatsu04} and Costanza~\cite{Costanza22} for surveys of {\small DLO} modeling techniques. In the robotics literature, sampling-based approaches are used to plan {\small DLO} steering paths~\cite{kavraki06}. 
Bretl~\cite{bretl_ijrr14} describes the {\small DLO}'s total elastic energy minimization as an optimal control problem, then uses the {\em adjoint equations} to design sample-based planners that steer flexible cables in {\small 2-D} and {\small 3-D} settings. Sintov~\cite{sintov20} built upon this work by pre-computing {\small DLO} equilibrium shapes for sampled costates, then determining dual arm steering paths using numerical tests to avoid self-intersections and obstacles.\\
\indent Local feedback control of {\small DLO} 
manipulation based on {\small MPC} schemes are described by Sintov~\cite{sintov21}, Yu~\cite{yu23} and Wang~\cite{wang22}. Navarro-Alarcon~\cite{Navarro-Alarcon16} and Lagneau~\cite{lagneau20} describe local shape control of {\small DLOs} using
the {\em deformation Jacobian,} which relates changes in the {\small DLO} shape to the robot grippers velocity. This paper focuses on single arm manipulation using Euler's elastica solutions for the {\small DLO} 
shapes~\cite{levyakov10}. 
The current paper extends  Gu's~\cite{GU23} vision based pipeline controller 
with a new {\em characterization} step, where  ResNet uses vision measurements to compute the current {\small DLO}'s elastica parameters that are fed to the feedback controller.\\
\indent \textbf{Paper contributions:} This paper proposes a two-layered approach for {\small DLO} placement on a~flat~surface~of~the~environ\-ment. Each layer comes with its own novel contributions. The high-level layer divides the {\small DLO} placement into three stages: {\em free transport} that brings the {\small DLO} tip into contact with the surface, {\em tip rolling} 
against the 
surface then {\em full rolling placement} of  the {\small DLO} under surface friction conditions. All stages are analytically characterized in~terms~of Euler's elastica solutions. The high-level planning of the robot gripper motion occurs in the space of elastica 
parameters defined for each placement stage. A key insight 
concerns~the~{\small DLO} curvature during placement. The {\small DLO}'s bending moment (which is proportional to its curvature) varies continuously along its length. 
As a result, the {\small DLO} start point of the portion aligned with the surface has {\em zero curvature.} 
This insight allows the high-level planner to analytically compute the {\small DLO}'s equilibrium shapes during placement.\\
\indent The paper's low-level layer forms a~pipeline controller consisting of two stages. The pipeline continuously estimates the flexible object's {\em medial axis} shape during placement. This medial axis forms the 
{\small DLO} model for the flexible object during placement. The {\small DLO} shape estimated in real time from an image taken by a~camera, then cropped and converted into the {\small DLO} synthetic image using \mbox{\small YOLO}v\mbox{\small 8}~\cite{YOLOv8}. The {\small DLO}'s synthetic image is the input to the 
pipeline second stage. This stage estimates by direct regression the elastica parameters of the {\small DLO} current shape using  ResNet50~\cite{ResNet}. The ResNet predicted elastica parameters are compared against the planned placement elastica parameters and fed into a~local feedback controller which commands the robot hand (Fig.~\ref{framework.fig}). Through simulations and comparative experiments,  the paper demonstrates the effectiveness of this two-layered {\small DLO} placement method.\\
\indent The paper is organized as follows. Section~II~\mbox{summarizes} Euler's elastica solutions for {\small DLO} equilibrium shapes. Section~III describes the three-stage placement approach together with 
the elastica parameters that determine the {\small DLO} shape in each stage. Section~IV describes the two-layered {\small DLO} placement scheme.  Section~V describes simulations statistics and experiments that compare 
semi-spatial placement of several mock-up fresh food items. Finally, Section~VI summarizes the results and suggests future research topics. 
\vspace{-.02in}  



\section*{\textbf{ II. DLO Equilibrium Shapes in Free Space}}
\vspace{-.02in}

\begin{figure}
\vspace{.1in}
\centerline{\includegraphics[width=0.5\textwidth]{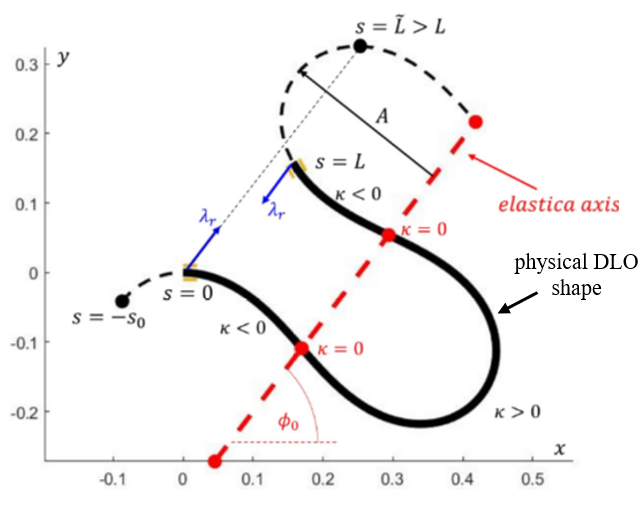}}
\vspace{-.18in}
\caption{Top view of full period elastica shape with the physical flexible cable of length $L$ embedded in its periodic elastica solution of full-period length~$\tilde{L}$.~The elastica axis with angle $\phi_0$ passes through the zero curvature points~and~is parallel to the opposing forces of magnitude~$\lambda_r$~applied~at~cable~endpoints.}
\label{full_period.fig}
\vspace{-.22in}
\end{figure}

\noindent This section reviews Euler's elastica solutions for {\small DLO} equilibrium shapes in two-dimensions.
The {\small DLO} is modeled as a non-stretchable elastic rod of length {\small $L$}  parameterized by arclength as $S(s) \!=\! (x(s),y(s),\phi(s))$ for $s \!\in\! [0,\mbox{\small $L$}]$. The {\small DLO} {\em state vector} consists of its $x,y$ coordinates and tangent direction $\phi$. The {\small DLO} curvature under arclength parametrization is given by $\kappa(s) \!=\! \tfrac{d}{ds}\phi(s)$. Using
curvature as control input, $u(s) \!=\! \kappa(s)$, the {\small DLO} shape is described by the {\small DLO} {\em control system}~\cite{bretl_ijrr14}:
\vspace{-.06in}
\begin{equation} \label{eq.dlo_sys}
  \frac{d}{ds}{S}(s)=
    \begin{pmatrix}
        \Dot{x}(s) \\
        \Dot{y}(s) \\
        \Dot{\phi}(s) 
    \end{pmatrix} =
    \begin{pmatrix}
        \cos\phi(s) \\
        \sin \phi(s) \\
        u(s)
    \end{pmatrix} \hspace{1.5em} \mbox{$s\in [0, L]$.}
\vspace{-.04in}
\end{equation}
\noindent One can now use optimal control tools to minimize the {\small DLO}'s total elastic energy 
\vspace{-.04in}
\[
    J = \mbox{\small $\frac{1}{2}$} EI \!\cdot\! \int_0^L u^2(s) ds 
\vspace{-.04in}
\]
\noindent under the {\small DLO} system constraints. Here            $E \!>\! 0$ is Young’s modulus of elasticity and $I \!>\! 0$ is the {\small DLO} cross-sectional second moment of inertia. The {\small DLO} {\em stiffness,} $EI$, is a known parameter. From calculus of variations, the {\em Hamiltonian}~\cite{opt_survey,ben-asher} of the {\small DLO} system \eqref{eq.dlo_sys} is given by
\vspace{-.04in}
\[ 
    H(S,\lambda, u) = \lambda_x \cdot \cos\phi + \lambda_y \cdot \sin\phi + \lambda_\phi \cdot u + \mbox{\small $\frac{1}{2}$} EI \cdot u^2
\vspace{-.04in}
\]
where $\lambda \!=\! (\lambda_x, \lambda_y, \lambda_\phi)$ are the {\em costate variables}. The costates correspond to internal force and bending moment along the {\small DLO}. The {\em adjoint equations}~\cite{pontryagin} determine the costate vector $\lambda(s)$ along energy extremal {\small DLO} shapes
\vspace{-.04in}
\begin{equation} \label{eq.adjoint}
  \frac{d}{ds}{\lambda}(s) \!= \! -\frac{\partial}{\partial S}  H(S,\lambda,u)\Rightarrow \!
        \begin{array}{lr}
        \Dot{\lambda}_x \!=\! 0 \\
        \Dot{\lambda}_y \!=\! 0 \\
        \Dot{\lambda}_\phi \!=\! \lambda_x \!\cdot\! \sin\phi(s) \!-\! \lambda_y \!\cdot\! \cos\phi(s)
        \end{array}
\end{equation}
\noindent Local minima shapes of the  {\small DLO}'s total elastic energy 
satisfy the additional condition
\vspace{-.01in}
\begin{equation}
    \frac{\partial}{\partial u}H\big(S,\lambda,u\big) = 0
\vspace{-.04in}
\end{equation}
\noindent which leads to Euler~Bernoulli bending law
\vspace{-.06in}
\begin{equation} \label{eq.bending_law}
    \lambda_\phi(s) + \mbox{\small $EI$} \cdot u(s) = 0 .
\vspace{-.06in}
\end{equation}
\noindent Using Eq.~\eqref{eq.adjoint}, $\lambda_x (s)$ and $\lambda_y(s)$~are constant along energy extremal {\small DLO} shapes. These constants define the 
parameter {\small$\lambda_r \!=\! \sqrt{\lambda_x^2 + \lambda_y^2}$}  which represents the~magnitude~of the opposing forces applied at the cable endpoints (blue arrows in Fig.~\ref{full_period.fig}), 
and the 
parameter $\phi_0$ which defines the angle of the {\em elastica axis} that passes through the zero curvature points of the periodic elastica solution (red axis in Fig.~\ref{full_period.fig}). 

Since the Hamiltonian of the {\small DLO} system has no explicit dependence on $s$, it remains constant along energy extremal {\small DLO} shapes, $H(s) \!=\! H^*$ for $s \!\in\! [0, \mbox{\small $L$}]$. 
Substituting $\lambda_\phi (s) \!=\! - \mbox{\small $EI$} \!\cdot\! u(s)$ from Eq.~\eqref{eq.bending_law} into $H(s)$ and replacing $u(s)$ by $\kappa(s)$: 
\vspace{-.12in}
\begin{equation} \label{eq.first}
    \lambda_r \cdot (\cos\phi(s)\cos\phi_0 + \sin\phi(s)\sin\phi_0)- \mbox{\small $\frac{1}{2}$} EI \!\cdot \kappa^2(s) = H^* .
\vspace{-.06in}
\end{equation}
\noindent Differentiating both sides w.r.t. $s$ gives
\vspace{-.1in}
\begin{equation} \label{eq.second}
    \lambda_r \!\cdot (-\sin\phi(s)\cos\phi_0+\cos\phi(s)\sin\phi_0)-EI \!\cdot \frac{d}{ds}\kappa(s)=0
\vspace{-.06in}
\end{equation}
\noindent where we canceled the common factor $\kappa(s) \!=\! \tfrac{d}{ds}\phi(s)$. Substituting the {\small DLO} system equations $\dot{x}(s) \!=\! \cos\phi(s)$ and $\dot{y}(s) \!=\! \sin\phi(s)$ into Eqs.~\eqref{eq.first}-\eqref{eq.second} gives
\vspace{-.06in}
\begin{equation*}
    \lambda_r \!\cdot\! R(\phi_0) \!\cdot\!
    \begin{pmatrix}
        \Dot{x}(s)\\
        \Dot{y}(s)
    \end{pmatrix}  = \begin{pmatrix}
        \frac{1}{2}EI\cdot \kappa ^2(s) \!+\! H^* \\
        EI\cdot \frac{d}{ds}\kappa(s)
    \end{pmatrix}    
\vspace{-.06in}
\end{equation*}
where $R(\phi_0) \!\!=\!{\footnotesize 
    \begin{bmatrix}
      \cos\phi_0 & \sin\phi_0 \\
      \sin\phi_0 & -\cos\phi_0 
    \end{bmatrix}}$.
\noindent Integrating both sides, $\int_{0}^{s} \Dot{x}(t)dt$ and $\int_{0}^{s} \Dot{y}(t) dt$, gives the  $(x,y)$ coordinates of the {\small DLO} in terms of its curvature~\cite{love}:
\vspace{-.06in}
\begin{equation} \label{eq.xy}
    \begin{pmatrix}
        x(s) \\
        y(s)
    \end{pmatrix} \!=\!  
    \begin{pmatrix}
        x(0) \\
        y(0)
    \end{pmatrix} \!+\! \frac{1}{\lambda_r} R(\phi_0) \!\cdot
    {\small  
    \begin{pmatrix}
        \frac{1}{2}EI\cdot\int_0^s \kappa ^2(\tau)d\tau+H^* \\[2pt]
        EI\cdot(\kappa(s)-\kappa(0))
    \end{pmatrix}}
\vspace{-.06in}
\end{equation}
\noindent where $s \!\in\! [0, \mbox{\small $L$}]$. Eq.~\eqref{eq.xy} describes the~{\small DLO}'s equilibrium shapes in terms of the 
control input $u(s) \!=\! \kappa(s)$. This paper focuses on {\em inflectional elastica}~\cite{love}
whose 
periodic solutions possess zero curvature (or inflectional) points (Fig.~\ref{full_period.fig}).


\textbf{The {\small DLO} elastica shape parameters:}
The {\small DLO} equilibrium shapes can be described by three elastica parameters (Fig.~\ref{full_period.fig}). These are the elliptic modulus parameter, $\mathrm{k}$, a~phase parameter~$s_0$ that measures the physical {\small DLO} start point, and the parameter~$\Tilde{L}$ that measures the full period length of the underlying elastica solution (Fig.~\ref{full_period.fig}). 
The {\small DLO} equilibrium shapes in two-dimensions are thus determined~by {\em six configuration parameters}: the {\small DLO}'s base frame position and orientation, $(x(0),y(0),\phi(0))$, and the elastica parameters $(\mathrm{k},s_0,\Tilde{L})$. 
\section*{\textbf{III. DLO Equilibrium Shapes in Contact with a Flat Surface of the Environment}}
\vspace{-.02in}

\noindent This section describes a three-stage scheme for {\small DLO}  placement on a planar surface of the environment.
To begin with,~it can be verified that gravity has only a~negligible effect on  
{\small DLO} equilibrium shapes provided that their length remains reasonably short, roughly up to  $50$~$\mathrm{cm}$ for strip like fresh food items~\cite{Levin&Grinberg&Rimon&Shapiro}[Appendix B]. Hence, gravity is present but its minor effect is ignored here.

\begin{figure}
\centerline{\includegraphics[width=0.5\textwidth]{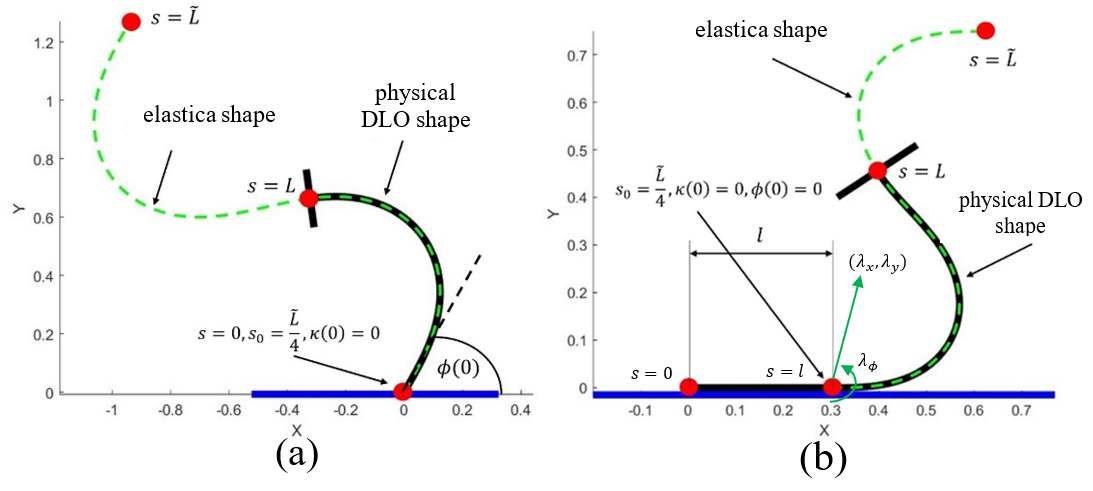}}
\vspace{-.08in}
\caption{(a) The {\small DLO} shape when the robot arm grasps the {\small DLO} at the red endpoint where $s = L$ while the {\small DLO} is supported by the surface at the tip where $s \!=\! 0$. When $s_0 \!=\! \tfrac{\Tilde{L}}{4}$, the {\small DLO} rotates clockwise 
around the surface attachment point. (b) The {\small DLO} shape when the robot arm grasps the {\small DLO} at the red endpoint where $s \!=\! L$ with a surface-contacting segment~of~length~$l$.}
\label{sub2_sub3.fig}
\vspace{-.2in}
\end{figure}

\textbf{DLO shape in contact with a planar surface:}
Consider the placement constraint $\mbox{\small $C$}(s) \!=\! y(s)- y_0$, where $y_0$ is the height of the plane on which the {\small DLO} should be placed (Fig.~\ref{sub2_sub3.fig}).
The elastic energy minimization problem now becomes
\vspace{-.06in}
\begin{mini}|s|
{u}{\mbox{\small $\frac{1}{2}$} EI \!\cdot\! \int_{0}^{s}{u^2(\tau)d\tau}}
{}{}
\addConstraint{
    \frac{d}{ds}{S(s)}=
    \begin{pmatrix}
   \cos\phi(s), \sin\phi(s), u(s)
    \end{pmatrix}} 
\addConstraint{and \ \ {y \geq y_0}.} 
\vspace{-.06in}
\end{mini}
\noindent When the surface constraint is active, $\mbox{\small $C$}(s) \!=\! 0$, the equations
\vspace{-.06in}
\begin{equation*}
        \begin{pmatrix}
        C(s) \\
        \frac{d}{ds}C(s) \\
    \end{pmatrix} \!=\! \begin{pmatrix}
        0 \\ 0
    \end{pmatrix} \!\!\Rightarrow \!\!\begin{pmatrix}
        y(s) \\ \sin\phi(s) 
        \end{pmatrix}\!=\! \begin{pmatrix}
            y_0\\0
    \end{pmatrix}
    \hspace{1.2em} \mbox{ $s\in[0,l]$} 
\vspace{-.06in}
\end{equation*}
\noindent where $l$ is the length of the
{\small DLO} segment that lies on the placement surface (Fig.~\ref{sub2_sub3.fig}(b)). The $(x,y,\phi)$ coordinates for the {\small DLO} surface contacting segment take the form
\vspace{-.06in}
\[ 
  N(s) = \left\{\begin{array}{lr}
         x(s) = x(0) + s  \\
         y(s) = y_0  \\
         \phi(s) = 0 \, \mbox{or}\, \pi
    \end{array}
    \right. \hspace{1.2em} \mbox{ $s\in[0,l]$}
\vspace{-.06in}
\]
\noindent  The {\em augmented Hamiltonian}~\cite{ben-asher,opt_survey} under the surface placement constraint is given by
\begin{equation}
   H(S,\lambda, u) \! = \!
        \left\{\begin{array}{lr}
        \lambda \!\cdot\! \dot{S} + \mbox{\small $\frac{1}{2}$} \mbox{\small $EI$} \!\cdot\! u^2, &  C(s) < 0   \\
        \lambda \!\cdot\! \dot{S} + \lambda_\mu \!\cdot\! C(s) + \mbox{\small $\frac{1}{2}$}  \mbox{\small $EI$} \!\cdot\!u^2 & C(s) \geq 0
    \end{array}
    \right.
\end{equation}
where $\lambda \in \mathbb{R}^3$ is the {\em costate vector} and $\lambda_\mu \leq 0$ is a~Lagrange multiplier of the surface-contacting constraint. 

Now consider the following observation concerning the {\small DLO} mechanics.
Euler-Bernoulli's bending law \eqref{eq.bending_law}  holds along the entire {\small DLO} length. The 
{\small DLO}'s curvature is thus proportional to its bending moment along its entire length.
The {\small DLO} curvature along its surface-contacting segment is identically zero. Since the bending moment $\lambda_\phi (s)$ varies {\em continuously} along the {\small DLO} entire length, the exit point from the surface-constraint segment at $s \!=\! l$ incurs zero bending moment and forms a~{\em zero curvature} (or inflection) point. 

In general, the elastica curvature equals to zero when $s \!+\! s_0 \!=\! \frac{\Tilde{L}}{4}$ or $s \!+\! s_0 \!=\! 
 \frac{3\Tilde{L}}{4}$ (Fig.~\ref{full_period.fig}).
The {\small DLO} shape during placement consists of two parts: a straight line from $s \!=\! 0$ to~$l$ aligned with the placement surface and the contact free portion from $s \!=\! l$ to~{\small $L$}~(Fig.~\ref{sub2_sub3.fig}(b)). The elastica parameter $s_0$ is set at the inflection point $\frac{\Tilde{L}}{4}$ when
the {\small DLO} has to be placed with {\em rightward} rolling (Fig.~\ref{sub2_sub3.fig}), and at the inflection point $\frac{3\Tilde{L}}{4}$ when
the {\small DLO} has to be placed with {\em leftward} rolling direction.

\begin{figure} 
\centerline{\includegraphics[width=0.5\textwidth]{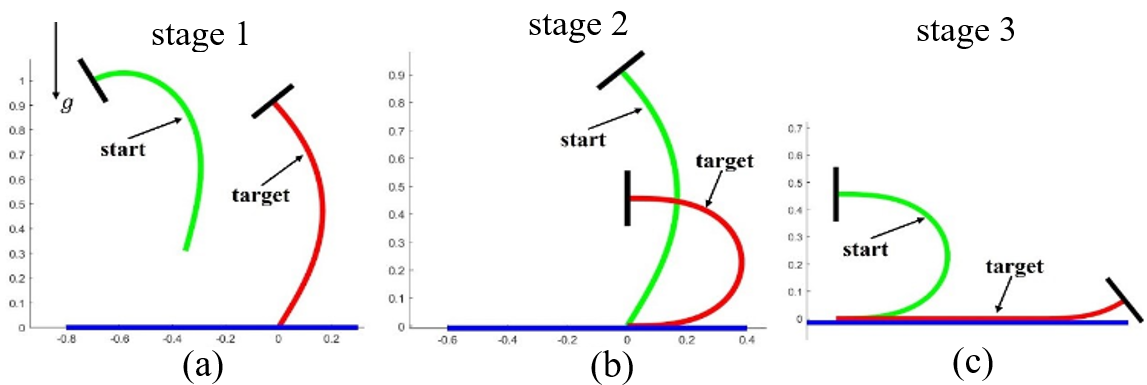}}
\vspace{-.06in}
\caption{(a) Stage I: contact free transport until the {\small DLO} tip touches the placement surface. (b) Stage II: non-slip rolling of the {\small DLO} tip against the surface until $\phi(0) \!=\! 0$. (c) Stage III: non-slip rolling placement of the entire {\small DLO} under surface friction conditions.}
\label{3_primitves.fig}
\vspace{-.2in}
\end{figure}

\textbf{{\small DLO} placement three configuration spaces:}
Each of the 
{\small DLO} placement stages uses a~different configuration space of 
parameters. In stage~I the robot arm transports the {\small DLO} suspended vertically under the influence of gravity until
its tip touches the placement surface 
(Fig.~\ref{3_primitves.fig}(a)). 
When stage~I is performed in two-dimensions, the {\small DLO} is steered in the six-dimensional configuration space consisting of 
the {\em DLO} base frame position and orientation {\small $(x(0),y(0),\phi(0))$} 
measured relative to the placement surface, then the contact free elastica parameters {\small$k,s_0,\Tilde{L}$}. When Stage~I is performed in a semi-spatial manner,
the {\small DLO} is steered in the eight-dimensional configuration space consisting of 
the {\small DLO} base frame position and orientation {\small $(x(0),y(0),z(0),\phi(0),\theta(0))$} where $\theta(0)$  is the base frame placement-plane rotation angle, then 
the contact free elastica parameters {\small $k,s_0,\Tilde{L}$}. The configuration space of Stage~I is thus $\mathcal{C}_1 \!=\!(x(0),y(0),\phi(0))\times(\mathrm{k},s_0,\Tilde{L}) \!\in\! \mathbb{R}^3\times\mathbb{R}^3\}$
or $\mathcal{C}_1 \!=\!\{(x(0),y(0),z(0),\phi(0),\theta(0))\times(\mathrm{k},s_0,\Tilde{L}) \in \mathbb{R}^5\times\mathbb{R}^3\}$. 

In stage~II the robot arm holds the {\small DLO} at its distal tip and executes non-slip rolling of the {\small DLO}'s tip while maintaining point contact with the surface until the {\small DLO} tangent at the tip aligns with the surface
at $\phi(0) \!=\! 0$
(Fig.~\ref{3_primitves.fig}(b)). In our practical setting the placement surface is a~packing tray placed on a~worktable. Hence, tip-tray as well as tray-table friction must be taken into account as discussed below. 
This paper assumes that the position of the  {\small DLO} tip touching the surface is known. The {\small DLO} shape is characterized by {\em zero curvature} at the tip,
as the contacting surface applies {\em zero  bending moment} on the {\small DLO}. Since the {\small DLO} has zero curvature at the tip, the elastica parameter $s_0$ (the elastica solution start point) equals~$\tfrac{\Tilde{L}}{4}$~or~$\tfrac{3\Tilde{L}}{4}$. 
In stage~II, three configuration parameters completely determine the {\small DLO} shape. This three-dimensional configuration space is defined as {\small$\mathcal{C}_2 \!=\! \{(\phi(0),\mathrm{k},\Tilde{L}) \in \mathbb{R}^3\}$}, where $\phi(0)$ is the tip tangent direction at the touch point, $k$ is the elastica modulus parameter and $\Tilde{L}$ is the full-period length elastica parameter. Note that $(x(0),y(0))$ (or $(x(0),y(0),z(0),\theta(0))$ in {\small 3-D}) as well as $s_0$ remain {\em constant} under the tip's non-slip constraint (Fig.~\ref{sub2_sub3.fig}(a)).\\
\indent In stage~III the robot arm executes non-slip rolling of the {\small DLO}'s contact free portion until it is fully placed on the surface 
(Fig.~\ref{3_primitves.fig}(c)).  In this stage the robot gripper holds the {\small DLO} at one endpoint while 
a~segment of length $l$ touches the surface followed by a~contact free elastica shape of length $L \!-\!l$ (Fig.~\ref{sub2_sub3.fig}(b)). 
In this stage, $s_0 \!=\! \tfrac{\Tilde{L}}{4}$ or $s_0 \!=\!\tfrac{3\Tilde{L}}{4} $ while the tip tangent is fixed at $\phi(0) \!=\! 0$. Hence, in stage~III three configuration parameters completely define the {\small DLO} shape, with one parameter changed from stage~II. This three-dimensional stage configuration space is defined as {\small$\mathcal{C}_3 \!=\! \{(l,\mathrm{k},\Tilde{L}) \in \mathbb{R}^3\}$}, where $l$ is the variable length of the segment contacing the surface, $k$ is the elastica modulus parameter and $\Tilde{L}$ is the full-period length elastica parameter. 

\begin{figure} 
\centerline{\includegraphics[width=0.5\textwidth]{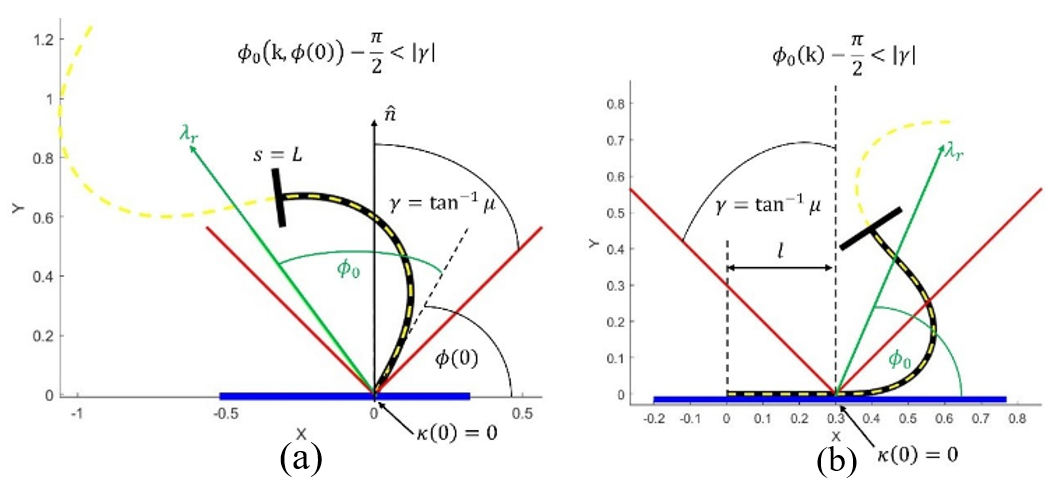}}
\vspace{-.1in}
\caption{The {\small DLO} non-slip constraint under Coulomb's  friction law in (a) stage II, and (b) stage III of the three-stage placement scheme.}
\label{friction_cone.fig}
\vspace{-.2in}
\end{figure}

\textbf{Non-slip placement constraint:}
To prevent the {\small DLO} from slipping on the tray the tray from slipping on the table during placement,
the {\small DLO}-tray and tray-table contact forces  must lie inside their {\em Coulomb friction conea}. 
These two non-slip constraints are lumped into a~single constraint using the coefficient of friction
$\mu \!=\! \min\{\mu_1,\mu_2\}$, where~$\mu_1$~is~the~coeff\-icient of friction between the {\small DLO} and the tray and $\mu_2$ is the coefficient of friction between the tray and the table. The composite non-slip constraint is met when the {\small DLO}-tray contact force lies inside the friction cone defined by $\mu$. Since the {\small DLO} curvature at the tip touching the surface is zero, the direction $\phi_0$ of the tip contact force 
can be computed as a function of the elastica modulus parameter $k$~\cite{Levin&Grinberg&Rimon&Shapiro}:
 \vspace{-.1in}
 \begin{equation} \label{eq.pm}
     \phi_0 = \phi(0) \pm \tan ^{-1}(1-2k^2)
     \vspace{-.1in}
 \end{equation}
where $\phi_0$ is also the direction of the elastica axis (Fig.~\ref{friction_cone.fig}). When $s_0 \!=\! \tfrac{\Tilde{L}}{4}$, a plus sign appears in Eq.~\eqref{eq.pm}. When $s_0 \!=\! \tfrac{3\Tilde{L}}{4}$, a minus sign appears in Eq.~~\eqref{eq.pm}. The friction cone constraint thus takes the form
\vspace{-.1in}
\begin{equation*}
    \phi_0 - \alpha(\hat{n}) < |\tan^{-1}(\mu)| \hspace{2em} \mu = \min\{\mu_1,\mu_2\}
    \vspace{-.08in}
\end{equation*}
where $\alpha(\hat{n})$ is the angle of the surface normal measured in the fixed world frame. For instance, $\alpha(\hat{n}) \!=\! \tfrac{\pi}{2}$ in Fig~\ref{friction_cone.fig}.

\section*{\textbf{IV. Local Shape Control Using Elastica Parameters and Machine Learning}}
\vspace{-.02in}

\noindent This section describes the {\small DLO}'s local shape control pipeline~(Fig.~\ref{framework.fig}). Since the elastica parameters charaterize the {\small DLO} shape, the control feedback computes a~multi-valued {\em inverse function} from the {\small DLO} measured image to the elastica shape parameters. Verification of the elastica parameters along the planned trajectory can give the current state estimation of the system and transfer the {\small DLO}'s observed shape to the controller. The controller can send a re-planning command that starts from the current state elastica parameters if the {\small DLO}'s shape error exceeeds a user-specified upper limit.

\textbf{Controlled placement scheme:} The {\small DLO} placement scheme planned a path. The algorithm utilized in this scheme can be found in our previous work~\cite{Levin&Grinberg&Rimon&Shapiro}, in this work the tray-contacting was added to the same scheme. Each node of the path is regarded as a ground truth. The low-level controller tries to minimize the {\em accuracy error} between the obtained {\small DLO} shape and the planned node. The accuracy error is the sum of the estimated shape error, the estimated elastica parameters error, and the estimated tangent error with different weights. The controller pipeline works using the following scheme: The input by the camera forms the current {\small DLO} shape frame, YOLOv8 crops the obtained frame from distractions, then extracts the DLO's medial axis~\cite{MedialAxis} (skeleton of the object). The {\small DLO}'s medial axis becomes the input to ResNet-50 network with three regression outputs, each one for a different elastica parameter. The output is the predicted elastica parameters. By knowing the predicted elastica parameters, the {\small DLO} shape is known. Since the inverse problem is {\em multi-valued}, ResNet is able to output several sets of candidate elastica parameters for the same measured {\small DLO} shape. Hence, the accuracy error between observed shapes determines if the real {\small DLO} shape is close to the planned shape. \\
\indent The controller pipeline scheme is summarized as Algorithm~\ref{alg:1}. The algorithm accepts as input the {\small DLO} length {\small $L$}, the {\small DLO} base frame $(x(0),y(0),\phi(0))$, and the {\small DLO} observed image from the camera. The algorithm computes the accuracy error {\small$||S-\hat{S}||$} and sends command to the robot arm according to the error value.

\begin{algorithm}[hbt!]
\textbf{Input:} $x(0),y(0),\phi(0)$, DLO length $L$, DLO image frame
\caption{Vision Based Local Shape Control}\label{alg:1}
\begin{algorithmic}[1]
    \State $cropped\_image \gets Crop(frame);$ 
    \State $rotate\_image \gets rotate(cropped\_image, \phi(0));$  
    \State $synthetical\_image \gets Skeleton(rotate\_image);$ 
    \State $(\Tilde{L},k,s_0) \gets Characterize(synthetical\_image);$
    \State $\hat{S} \gets Get\_shape(x(0),y(0),\phi(0),\Tilde{L},k,s_0);$
    \If {$||S-\hat{S}||\leq\varepsilon$}
       \State $Continue;$
    \Else
        \State $Recovery;$
    \EndIf
\end{algorithmic}
\end{algorithm}

\begin{figure} 
\centerline{\includegraphics[width=0.5\textwidth]{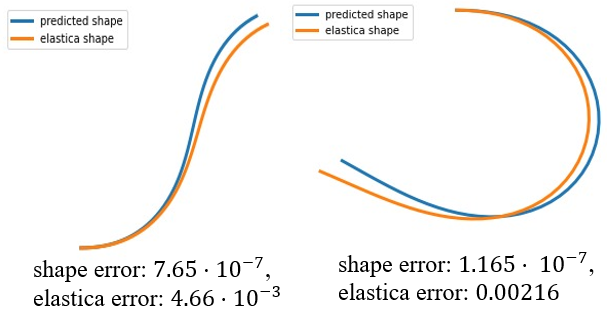}}
\caption{Evaluation methods of ResNet-50 are the shape error and the elastica parameters error: comparison between the ground truth DLO (orange) and predicted DLO (blue). The shape error is the distance between each point $(x(s),y(s))$ for $s \in [0, L]$ on each line. The elastica error is the MSE metric between the $[\Tilde{L},s_0.k]$ labels from the planer and the prediction from the ResNet-50 model. It can be inferred that the ResNet-50 model's prediction is good because the error is low.}
\label{preformens_testset.fig}
\vspace{-.2in}
\end{figure}

\textbf{Dataset collection and generation:}
\noindent To train the proposed two ML models, multiple datasets were utilized. The first synthetic dataset was generated using iterative for-loops over the elastica parameters to capture their behavior. Each iteration step was determined based on the search step of the high-level planning algorithm (the iteration steps are detailed in the experiments section). This dataset aims to train the ResNet-50 learning model to predict the elastica parameters. The second synthetic dataset has the same purpose as the first dataset. However, the data underwent image processing to reduce noise and minimize interference in the ResNet-50 model learning process. The third dataset is an experimental one. This dataset aims to train the YOLOv8 model to leave the observed frame only with the DLO. The dataset was obtained through video recordings of DLOs using the laboratory camera (model specified in the experiments section), capturing frame-by-frame the obtained DLO shape. This multi-source dataset approach enabled comprehensive training of the controller pipeline across different scenarios.\\ 
\begin{figure}
\centerline{\includegraphics[width=0.48\textwidth]{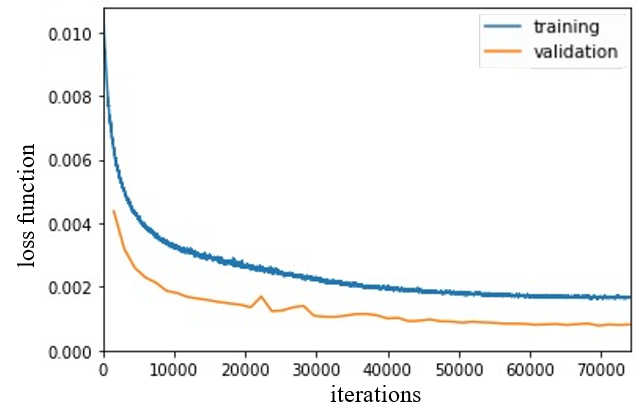}}
\caption{ The loss optimization process for the ResNet-50 model.}
\label{preformens_loss.fig}
\end{figure}
\indent \textbf{Training and evaluation methodology:} 
\noindent The controller pipeline training process was conducted in multiple stages: the initial stage was to train the ResNet-50 regression model to predict the elastica parameters using the first synthetic dataset. This stage tested three loss optimization functions, L1, MSE, and Huber, to optimize the ResNet-50 model predictions. The evaluation was performed on individual shapes using a structured dataset split into training, validation, and test sets. The accuracy metric {\small$||S-\hat{S}||$}, was checked for each epoch. Through this analysis, the identified error threshold required to achieve an accuracy level exceeding 90\%. The results indicated that MSE and Huber loss functions provided the best performance (Fig.~\ref{preformens_loss.fig}). The  {\em Huber loss function}~\cite{Huberloss} takes the form
\vspace{-.15in}
\begin{equation*}
    Loss_{h} = \begin{cases}
        0.5 (x_n - y_n)^2, & \text{if } |x_n - y_n| < \delta \\
        \delta \cdot (|x_n - y_n| - 0.5 \cdot \delta) & \text{otherwise }
        \end{cases}
\vspace{-.12in}
\end{equation*}
\noindent where {\small$x_n$} is the vector of ground truth elastica parameters~and {\small$y_n$} is the vector of predicted elastica parameters, $\delta$ is a~numerical value that defines the boundary where Huber's loss function transitions from quadratic to linear. The second training stage was to try to optimize ResNet-50 prediction using the second dataset, which led to an improvement in the average accuracy error in the total trajectory. These two stages form the {\small DLO} {\em characterization model}. Examples comparing the {\small DLO} shape and elastica parameters errors from the test set are shown in Fig.~\ref{preformens_testset.fig}.\\
\indent During the final training stage {\small YOLO}v{\small 8} model was trained to crop the current observed {\small DLO} frame. Using the experimental dataset to generate focused, cropped images. The {\small YOLO}v{\small 8} model was trained on a minimal dataset and has done the {\small DLO} isolation. The cropped images were then passed to the controller pipeline for further refinement. The {\small YOLO}v{\small 8} model mask cannot be considered a characterization of the {\small DLO} shape because the cropped image has many inaccuracies, which the estimation of the elastica parameters corrects and guarantees the {\small DLO} shape.

\section*{\textbf{V. Representative Simulations and Experiments}}
\vspace{-.02in}

\noindent This section describes representative simulations followed by
{\small DLO} placement experiments. The 
placement planning code will be posted at the Github~\cite{Git_grinberg&levin}.

\begin{figure} 
\centerline{\includegraphics[width=0.5\textwidth]{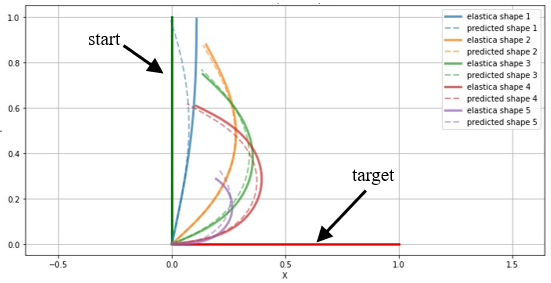}}
\vspace{-.08in}
\caption{ Five {\small DLO} frames sampled during placement. The green vertical line is the {\small DLO} start position, the horizontal red line is the {\small DLO} target placement. The colors distinguish between each sampled frame. In each frame the bold curve is the elastica planned shape and the dashed curve is the ResNet predicted {\small DLO} shape. The surface attachment point is located at $(0,0)$.}
\vspace{-.15in}
\label{sim_one.fig}
\end{figure}

\begin{figure}
\centerline{\includegraphics[width=0.5\textwidth]{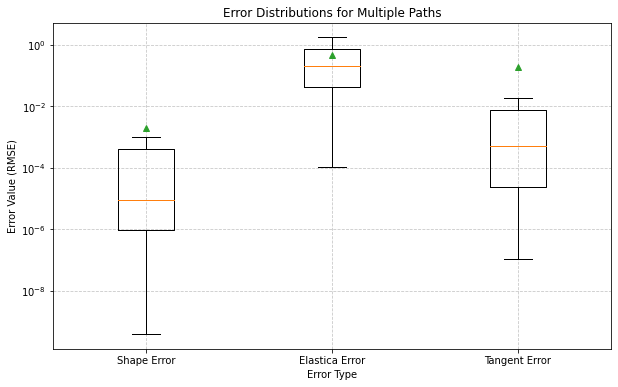}}
\caption{ The three error distributions: shape error, elastica error, and tangent error. The orange line is the median of each error, and the green triangle is the mean. It can be understood from the box-plot that more than 90\% of the predicted shapes are good. It means the controller works well. The error value axis is in log scale.}
\label{sim_two.fig}
\vspace{-.15in}
\end{figure}

\textbf{Simulation results:}
The simulations are aimed at evaluating the low-level controller on synthetic {\small DLO} placement paths and to demonstrate the DLO placement scheme. The placement scheme was implemented in Python and run on DELL Precision 7865 with Ubuntu 22.04 desktop, AMD Threadripper PRO 5945WX Processor, 32GB RAM, 16GB GPU Quadro RTX a4000. The placement scheme received as input the base frame position $x(0),y(0),\phi$ and elastica parameters $\Tilde{L},s_0, k $ of the DLO start and target positions. The resolution of the configuration parameters: $\Delta x(0) = \Delta y(0)=\Delta z(0)=0.01\cdot L$~$\mathrm{[m]}$, $\Delta \phi(0)=\Delta \theta=2^{\circ}$, $\Delta \Tilde{L} = 0.02L$, $\Delta k = 0.005$, $\Delta l=0.05\cdot L$, and $s_0 =\frac{\Tilde{L}}{4}$. The sampling rate is 7 [fps]. The output is the observed DLO shape projection. This shape was inserted into the controller pipeline. The section describes three ranking methods of the controller performance: the shape error, the elastica parameters error, and the tangent error. The errors are between the observed DLO configuration and the planned configuration in each sample frame on the manipulation trajectory. The results are shown in Fig.~\ref{sim_one.fig}. Five sampled frames simulated and obtained path are compared in Fig.~\ref{sim_one.fig}. A multiple-path checking was made on 22 paths including 1015 shapes shown in Fig~\ref{sim_two.fig}. Each box represents an error type distribution based on the mean and std. These distributions are used to determine the accuracy error threshold $\varepsilon$. The controller will send a re-planning message when the computed accuracy error exceeds this value. The shape error is {\small $0.0020 \pm 0.0061 \ [m]$}, the elastica error is {\small $0.4766 \pm 0.6288$}, and the tangent error is {\small $0.1873 \pm 0.8721 \ [rad]$}. All errors are shown by the mean $\pm$ standard deviation. Because the problem of the elastica parameters is a multi-valued function, it can be seen that the elastica parameters mean error is close to the median (Fig.~\ref{sim_two.fig}). That is why the elastica parameters have a low weight relative to the other errors.     

\begin{figure}
\centerline{\includegraphics[width=0.49\textwidth]{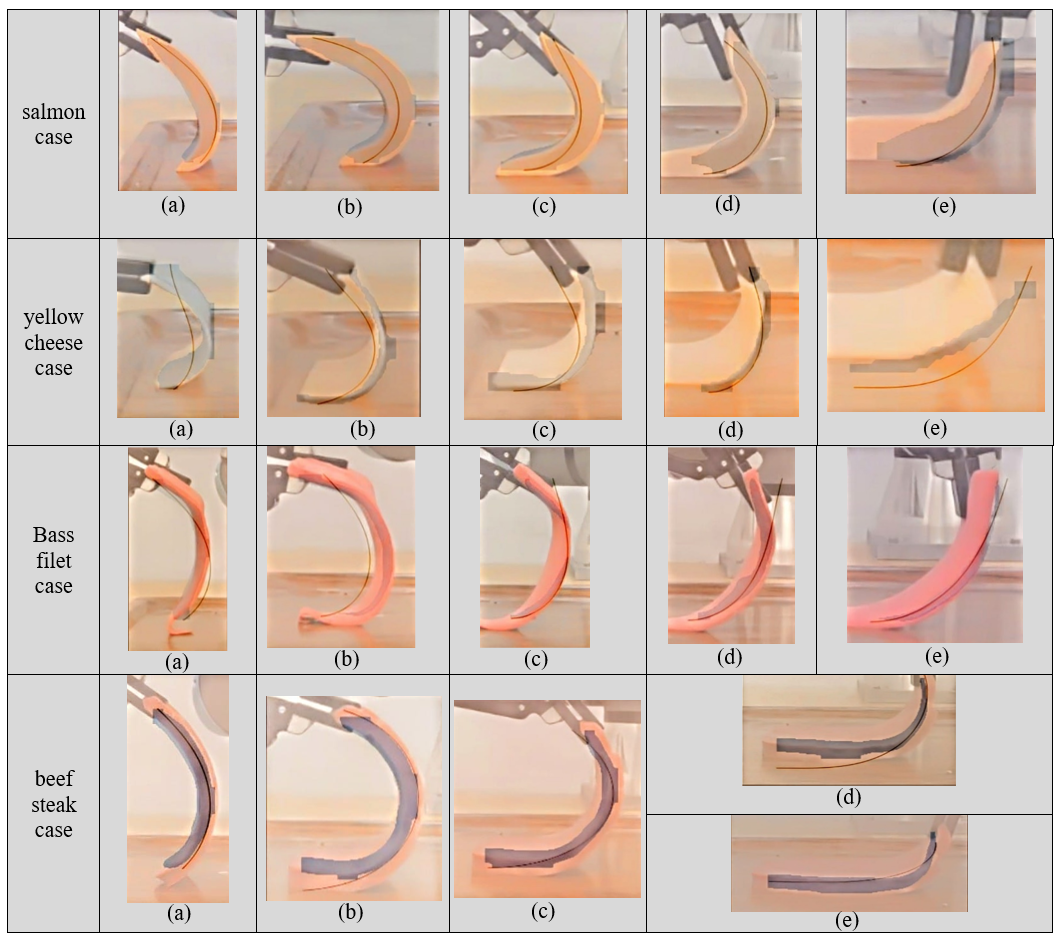}}
\caption{(a)-(e) Four different cases of the two-layered {\small DLO} placement experiments. Each image shows a particular frame of the {\small DLO} placement process. Each image describes the three layers of the controller pipeline. The first layer (background) is the crop stage, the second layer (blue polygon) is the image synthetic stage, the third layer (black curve) shows the observed {\small DLO} shape computed by ResNet-50's predicted elastica parameters.}
\label{cases.fig}
\vspace{-.25in}
\end{figure}

\begin{figure}
\centerline{\includegraphics[width=0.52\textwidth]{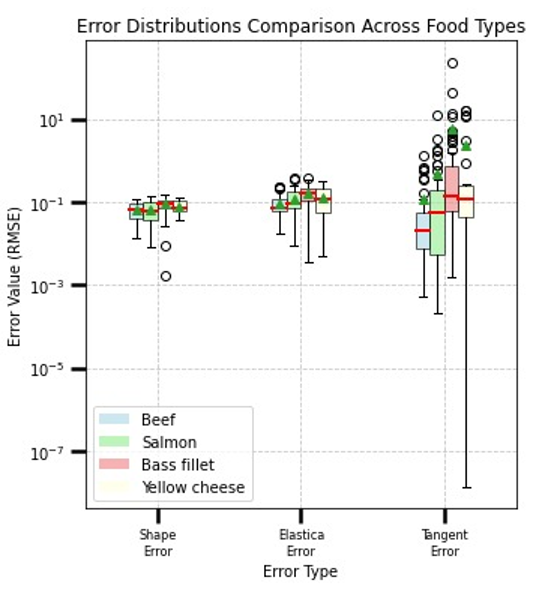}}
\caption{Three error distributions in real use-cases. In the shape and elastica errors, the median (red lines) and mean (green triangle) error of observed shapes and the elastica parameters are similar and very close in the different cases. Thus indicates that there is no effect of the object type and it can be assumed that they are all DLOs. In contrast, the tangent error is influenced by the type of object thickness.}
\label{real_errors.fig}
\vspace{-.15in}
\end{figure}

\indent \textbf{Real world experiments:}
 The proposed framework has been validated through experiments. Four silicon mock-up objects were prepared: steak, salmon, bass fillet and yellow cheese. The experiment setup is shown in Fig.~\ref{framework.fig}. A~UR5e robot arm with a~Robotiq 2F-85 gripper performs the {\small DLOs} placement on a tray supported by a worktable. A Logi-1080HD web camera is utilized to observe the {\small DLO} shape. The camera is exposed to the profile of the {\small DLO}. The three verification methods proposed were: execution time, task success, and small accuracy error. A comparison was made between four different mock-up silicon objects. The task is to place the object comfortably on a tray without slipping or wrinkling, which can damage the {\small DLO}. In Fig.~\ref{cases.fig} for all cases the task was successful. The two-layered placement average execution time was $55$~seconds with standard deviation of $5$~seconds. There is a correlation between task execution time and object length {\small $L$}, with smaller objects taking less time to place. In that manner of the accuracy error, the shape error is {\small $0.00752 \pm 0.00306 \ [m]$}, the elastica error is {\small $0.1271 \pm 0.0774$}. Those errors in the steak and the salmon cases are better than the cheese and bass fillet objects (Fig.~\ref{real_errors.fig}). The tangent error has a large mean and standard deviation. This can be explained  as sensitivity  of the tangent error to the {\small DLO}  thickness. As the cheese and the bass fillet objects are thinner than the steak and salmon objects. The controller pipeline thus works better with the salmon and beef steak objects.

\section*{\textbf {VI. CONCLUSION}}
\vspace{-.05in}

\noindent The paper described a two-layered {\small DLO} surface placement approach. The high-level layer is a three-stage placement scheme based on Euler's elastica solutions. The low-level layer is a~visual perception controller monitoring execution of the robot arm placement process. The high-level layer uses three stages to plan the {\small DLO} placement. First free transport in a six-dimensional configuration space (eight-dimensional in the case of semi-spatial placement), then two types of three-dimensional configuration spaces of the elastica parameters during {\small DLO}  during tip rolling and full placement in contact with the planar surface. All placement stages are performed under non-slip conditions in a locally stable through local minima of the {\small DLO}'s elastic energy and in non self-intersecting manner. Practical realization of the {\small DLO} planned placement path is monitored by a low-level perception
controller that uses {\em ML} pipeline to estimate  the accuracy error between the observed and planned {\small DLO} shapes during task execution in the presence of modeling and placement errors.\\  
\indent Future research will consider several topics. The high-level planner will be augmented by computation of the {\small DLO} equilibrium shapes under {\em elastic and gravitational} energies. This extension will require numerical solution of the {\small DLO}'s adjoint equations rather than the closed form elastica solutions.  In addition, the reverse process of this paper's placement scheme could be adapted for {\small DLO} pick-up from a~surface such as a conveyor belt. 
At the low-level control, 
under this paper's method when the controller gets a high accuracy error, it sends a command of recovery and  re-plans the {\small DLO}'s placement path from the current obsreved state. By adding a~{\em Kalman filter},  prediction of the {\small DLO} shape and position can be made before the manipulation starts. The controller will then be able to estimate the accuracy error and react before any physical movement of the robot arm begins. Finally, reinforcement and imitation learning techniques can learn a policy and rewards based on this paper's accuracy error, then improve on the ResNet elastica parameters computation used by this paper.  


\bibliographystyle{IEEEtran}

\begin{bibliography}{books,elonbib,elon}
\end{bibliography}

\end{document}